\documentclass[letterpaper]{article} % DO NOT CHANGE THIS
\usepackage[preprint]{aaai2027}
\usepackage[hyphens]{url} % DO NOT CHANGE THIS
\usepackage{graphicx} % DO NOT CHANGE THIS
\usepackage{natbib} % DO NOT CHANGE THIS AND DO NOT ADD ANY OPTIONS TO IT
\usepackage{caption} % DO NOT CHANGE THIS AND DO NOT ADD ANY OPTIONS TO IT
\usepackage{booktabs}
\usepackage{multirow}
\usepackage{amsmath}
\usepackage{amssymb}
\newcommand{\system}{\textsc{MemTxn}}
\newcommand{\twopanellabels}[2]{%
  \noindent
  \makebox[\columnwidth][c]{%
    \parbox[t]{0.48\columnwidth}{\centering\small (a)~#1}%
    \hfill
    \parbox[t]{0.48\columnwidth}{\centering\small (b)~#2}%
  }%
}

\title{MemTxn: A Transaction Boundary for Source-Supported Updates and Complete-State Recovery in Agent Memory}
\author{Hanshuai Cui\textsuperscript{2,1}, Zhiqing Tang\textsuperscript{1}, Zhi Yao\textsuperscript{2,1}, Fanshuai Meng\textsuperscript{1}, Qianli Ma\textsuperscript{1}, Weijia Jia\textsuperscript{1}}
\affiliations{\textsuperscript{1}Institute of Artificial Intelligence and Future Networks, Beijing Normal University, Zhuhai 519087, China\\
\textsuperscript{2}School of Artificial Intelligence, Beijing Normal University, Beijing 100875, China}

\begin{document}
\maketitle

\begin{abstract}
Persistent memory lets long-running large language model agents reuse information across sessions and tasks. Yet errors in writable memory can persist and corrupt future behavior. Existing systems improve storage and retrieval, but they do not provide a transaction boundary for reliable updates and recovery. We therefore propose \system{}, a governance layer outside the answer model. \system{} verifies whether an update is supported by its source. It also selects the visible version when facts conflict and restores the application-visible state after a fault. The system uses Ordered PatchTest to validate writes, a Temporal Resolver to select versions, and a durable snapshot journal to recover state. On an item-disjoint audit, \system{} accepts all 60 supported originals and rejects all 179 hard negatives. Under persistent multi-key faults on LongMemEval-S and LoCoMo states, it restores the complete declared active map without knowing the actual physical write set. On MemoryAgentBench FactConsolidation, \system{} achieves the highest average F1 across all twelve answer-model configurations. It outperforms Dense by $17.06$--$24.07$ points in five representative settings.
\end{abstract}

\section{Introduction}

\begin{figure}[t]
\centering
\includegraphics[width=\columnwidth]{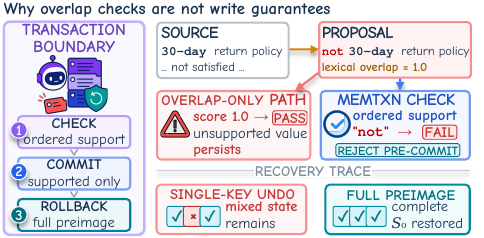}
\caption{MemTxn rejects unsupported edits before commit and restores the full application-state preimage after a fault.}
\label{fig:teaser}
\end{figure}

Long-running assistants and memory-augmented agents increasingly persist conversational facts, preferences, and task state for decisions made hours or sessions later~\cite{park2023generative,packer2023memgpt,mem02025}. Writable memory can make unsupported extractions durable and expose the wrong version when facts conflict. Partial writes can also corrupt future tasks~\cite{xiong2026memorymanagement}. The central problem is therefore no longer only \emph{which memory is relevant}. A memory system must decide which proposed state may commit and which conflicting version should be visible. It must also identify the prior application state to restore after failure. Figure~\ref{fig:teaser} summarizes \system{}'s write and recovery paths. In the write trace, a negated return-policy update from a held-out LongMemEval-S item~\cite{longmemeval2024} retains lexical support 1.0 but fails ordered support; \system{} rejects it and commits only the supported value. In the recovery trace, \system{} restores the complete application-visible preimage after a persistent fault, whereas single-key undo can leave a mixed state.

Earlier systems retrieve and reflect over episodic experience, while MemGPT manages short- and long-term memory tiers~\cite{park2023generative,shinn2023reflexion,packer2023memgpt}. Retrieval-augmented generation and dense passage retrieval rank external evidence for downstream generation~\cite{lewis2020rag,karpukhin2020dpr}. Recent systems move beyond passive retrieval: they extract persistent user memories, organize structured stores, compress memory, model temporal relations, and learn memory operations~\cite{mem02025,amem2025,memoryos2025,memos2025,lightmem2025,zep2025,yan2025memoryr1}. Together these advances make persistent memory more editable and autonomous.

Yet relevance, recency, and learned memory actions do not constitute a commit contract. They neither require source support for a proposed value nor distinguish answer-time fallback from rejection of a durable write. They also do not specify the complete application-visible state to restore when an invariant violation remains after a database reopen. Our key observation is that an external-memory update can be governed as an \emph{application-level transaction}. Evidence support governs admission, while an explicit conflict contract selects the committed version visible to an answer. A persisted preimage enables recovery. Storage atomicity can make a physical write all-or-nothing, but it cannot enforce these semantic contracts.

Building such a transaction boundary for writable agent memory raises two challenges.  \emph{(C1) How to decide which proposed state may commit and which conflicting version should be visible without answer supervision?}  Subtle extraction errors can preserve nearly all source words while changing their order or polarity, whereas relevance or recency alone cannot establish source support.  Consulting gold answers, benchmark labels, or future queries would instead leak evaluation signals into persistent state.  \emph{(C2) How to restore the complete intended application state after a durable multi-key fault without knowing the actual physical write set?}  A single logical memory update can modify versions, pointers, fact keys, and event records across several physical keys.  A single-key baseline can therefore leave a mixed state that survives reopen, while storage atomicity alone cannot identify the application-level preimage to restore.

We introduce \system{}, an answer-model-external governance layer with three governance paths.  Ordered PatchTest activates only proposals satisfying the declared ordered lexical support contract against a cited span.  A deterministic conflict trigger and Temporal Resolver select the version visible to the answer model under a declared chronology contract.  A durable snapshot journal restores the audited active map after an invariant failure.  In the item-disjoint audit, the frozen gate accepts all 60 supported originals and rejects all 179 audited hard negatives.  Across controlled faults on LongMemEval-S and LoCoMo states, \system{} restores the complete active map without the physical write set, while the single-key baseline leaves mixed states.  On MemoryAgentBench FactConsolidation~\cite{memoryagentbench2026}, \system{} achieves the highest average F1 across all twelve answer-model configurations. In five representative settings, it outperforms Dense by 17.06--24.07 points. We make the following contributions.
\begin{itemize}
    \item We identify the missing \emph{transaction boundary} in writable agent memory and formalize separate contracts for source-supported update admission, conflict-conditioned visibility, and complete-state recovery.
    \item We develop \system{} with answer-independent admission, temporal resolution, and invariant-checked recovery while keeping rejection, fallback, and rollback observable.
    \item We validate \system{} on source-support, recovery, and FactConsolidation evaluations, achieving exact correctness in both audits and the highest average F1 across all twelve answer-model configurations.
\end{itemize}

\section{Related Work}

\paragraph{Writable agent memory.}
Generative Agents~\cite{park2023generative} retrieves and reflects over episodic memories, while MemGPT/Letta~\cite{packer2023memgpt} manages a memory hierarchy.  Voyager~\cite{wang2023voyager} and Reflexion~\cite{shinn2023reflexion} retain reusable skills or verbalized experience.  Recent surveys organize agent memory by cognitive role and by its progression from storage to experience~\cite{wu2025humanmemory,luo2026storage}.  Mem0~\cite{mem02025} supports persistent user memories; A-MEM~\cite{amem2025} constructs autonomous notes; MemoryOS~\cite{memoryos2025} and MemOS~\cite{memos2025} organize hierarchical stores; LightMem~\cite{lightmem2025} compresses memory; LangMem~\cite{langmem2025} provides long-term memory tools; and Zep~\cite{zep2025} builds temporal graphs.  Empirical evidence nevertheless shows that retained experience can propagate errors or become misaligned with later tasks~\cite{xiong2026memorymanagement}. The Agent-Memory Protocol~\cite{wu2026amp} focuses on protecting identifiers at the user boundary.  \system{} addresses the orthogonal transaction boundary for write activation, conflict visibility, and complete application-state recovery.

\paragraph{Retrieval and long-memory evaluation.}
RAG~\cite{lewis2020rag} and dense passage retrieval~\cite{karpukhin2020dpr} expose external knowledge to a generator.  BGE-M3~\cite{chen2024bgem3} provides multilingual dense retrieval.  LongMemEval~\cite{longmemeval2024}, LoCoMo~\cite{locomo2024}, MemoryAgentBench~\cite{memoryagentbench2026}, BEAM~\cite{beam2026}, MemBench~\cite{membench2025}, Mem2ActBench~\cite{shen2026mem2actbench}, and RealMem~\cite{bian2026realmem} measure complementary conversational, incremental, tool-use, and project-state abilities.  These benchmarks define evaluation units rather than memory transaction protocols.  Our natural and controlled studies provide complementary evidence at their respective units.

\paragraph{Editing and regression safety.}
KnowledgeEditor~\cite{decao2021knowledgeeditor}, MEND~\cite{mitchell2022mend}, ROME~\cite{meng2022rome}, MEMIT~\cite{meng2023memit}, WISE~\cite{wang2024wise}, and AlphaEdit~\cite{fang2025alphaedit} modify parametric knowledge.  KnowEdit~\cite{zhang2024knowedit} surveys and benchmarks that setting, while Memory-R1~\cite{yan2025memoryr1} learns operations for external stores.  \system{} leaves model weights unchanged. Write-ahead logging, partial rollback, checkpoints, and atomic commit provide the storage foundation~\cite{mohan1992aries,sqlitewal}.

\system{} lifts transaction semantics to agent memory by making cited-span write support, answer-context selection, recovery intent, and application-level restoration independently observable.  Storage atomicity can faithfully persist a semantically invalid state.  MemTxn governs that higher-level state and supplies the compensating action required after a declared abort or memory-invariant failure.  Tool-using agent studies illustrate realistic state-changing settings~\cite{schick2023toolformer,yao2024taubench}.  Safety benchmarks further motivate persistent-state checks because a write can outlive the interaction that caused it~\cite{ruan2023toolemu,agentdojo2024,zhang2025asb}.

\begin{figure*}[t]
\centering
\includegraphics[width=\textwidth]{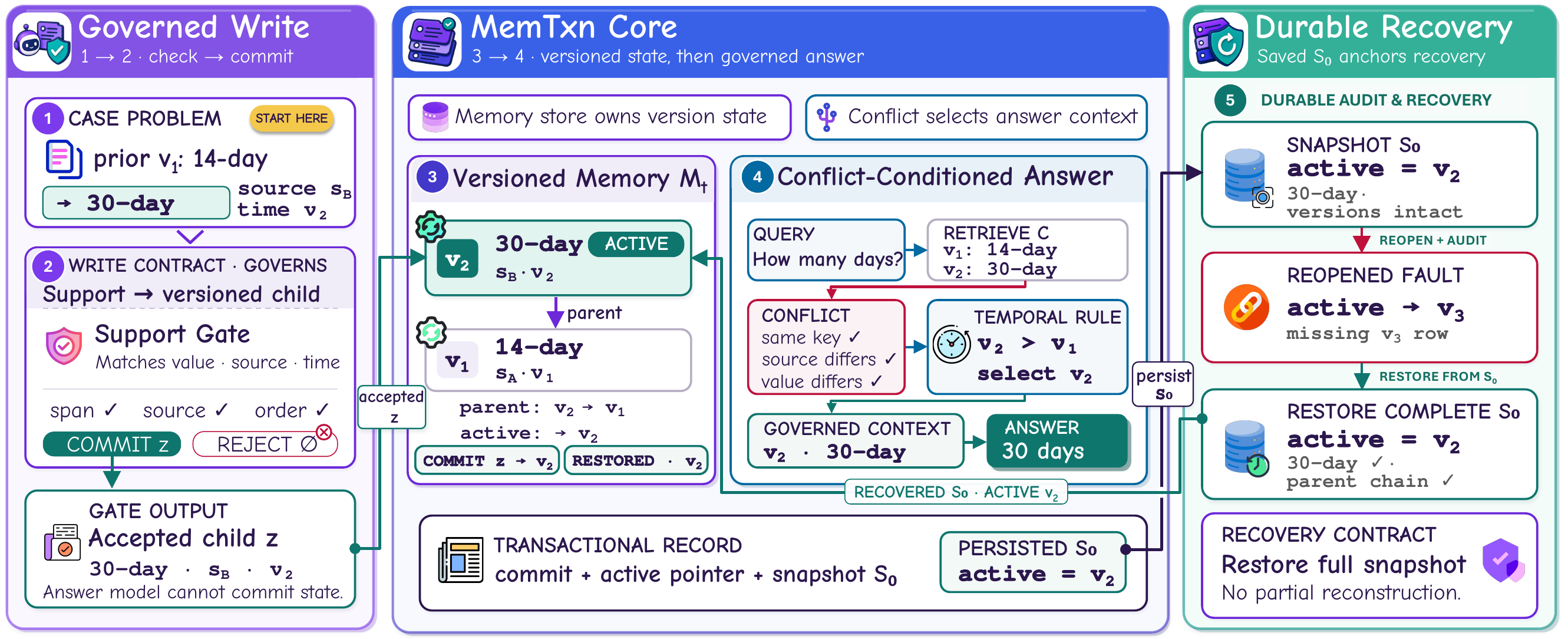}
\caption{MemTxn workflow on the return-policy example. The write lane governs activation, the answer lane resolves conflicts, and the audit lane restores state.}
\label{fig:lifecycle}
\end{figure*}

\section{Method}

\subsection{Overview}

\system{} is an answer-model-external governance layer with three separate paths: source-supported update admission, conflict-conditioned answer-context selection, and durable recovery of the complete application-visible state.  This separation prevents an answer-time route decision from silently becoming a write decision. It also prevents a storage-level repair from being mistaken for restoration of the intended application state.

Figure~\ref{fig:lifecycle} reads from left to right and follows the same return-policy key through five numbered steps. In Steps 1--2, the held-out source supports \texttt{30-day} against the illustrative prior $v_1=\texttt{14-day}$. After the metadata checks and Ordered PatchTest pass, the store creates $v_2$, links it to $v_1$, and makes $v_2$ active. In Steps 3--4, a conflict-conditioned query retrieves both versions, and the Temporal Resolver selects $v_2$ under the declared chronology. In Step 5, an injected fault survives a database reopen: the active pointer refers to $v_3$, but the $v_3$ row is missing. Recovery restores the saved application-visible preimage $A_0=\Pi_{\rm app}(S_0)$, leaving $v_2$ active.

The figure also distinguishes state-changing decisions from answer-time routing. Rejection leaves persistent state unchanged, while a governed answer uses the selected context without authorizing a write. Write activation is answer-independent: neither labels nor a future query can rescue an unsupported proposal. Recovery is durability-aware: rollback is counted only after the fault survives reopen and restoration re-establishes the complete saved active map. Both admission and recovery remain observable outside the answer model.

\subsection{System model and trust boundary}

At logical time $t$, persistent memory is $S_t=(\mathcal V_t,A_t,\mathcal J_t)$, where $\mathcal V_t$ is an append-only set of versions, $A_t:\kappa\mapsto\mathrm{vid}$ is the active-version map, and $\mathcal J_t$ contains durable state events and recovery intents.  We write $\Pi_{\rm app}(S_t)=A_t$ for the application-visible active-state projection; repair may append to $\mathcal V_t$ and $\mathcal J_t$ without changing the recovered projection.

An upstream extractor proposes $q=(\mathrm{id},\kappa,u,r,v,e,s,\nu)$, where $\kappa$ is a normalized key, $u$ and $r$ are subject and relation, $v$ is the value, $e$ is a cited evidence span, $s$ is an opaque source ID, and $\nu$ is chronology metadata.  The store augments an accepted proposal into $z=(\mathrm{vid},q,\rho,\sigma,t_c)$ with version ID, parent pointer, lifecycle status, and commit time.  Parent pointers and lifecycle status are not extractor outputs; frozen regression tasks remain evaluator-side.

A governed update is an application-level transaction $\tau=(q,S_0,K_\tau,\sigma_\tau)$ with a pre-update state, declared logical scope, and a status in \textsc{Proposed}, \textsc{Rejected}, \textsc{Active}, \textsc{Superseded}, or \textsc{RolledBack}. Its recovery intent persists the application-visible preimage $A_0=\Pi_{\rm app}(S_0)$ rather than the entire physical state. Its interface intentionally omits the physical write set.  SQLite supplies atomic writes; \system{} supplies semantic admission, visibility, and compensating recovery.

The proposer, candidate value, and writable child are untrusted. SQLite, the persisted intent, and the saved active-map snapshot form the recovery trusted base.  Ordered PatchTest receives only proposal fields and the raw source referenced by the opaque source ID. Gold answers, benchmark types, session IDs, safety labels, and fault labels enter only after runtime decisions.  The boundary establishes cited-source consistency, declared-chronology visibility, and restoration from an intact intent. It does not guarantee source truth, semantic role binding, concurrency, repeated-fault tolerance, or recovery from physical loss.

\subsection{Source-supported update admission}

Let $T(x)$ be the normalized content-token sequence of text $x$, $\preceq$ the ordered-subsequence relation, and $F(q)$ an indicator that key, value, source ID, chronology, and evidence are all present and $T(v)\neq\emptyset$.  Let $D_s$ be the source text referenced by opaque ID $s$, and let $Q(e,D_s)$ indicate that $\operatorname{norm}(e)$ is a substring of $\operatorname{norm}(D_s)$.  Lexical baselines accept when metadata and source checks pass and the fraction of candidate-token occurrences covered by the cited evidence reaches the fixed or calibrated threshold $\theta$ in Table~\ref{tab:audits}.  The frozen Ordered PatchTest instead uses
\begin{equation}
\label{eq:ordered-support}
\operatorname{Support}_{\rm ord}(q)=F(q)\land Q(e,D_s)
\land[T(v)\preceq T(e)].
\end{equation}
Equation~\ref{eq:ordered-support} requires every value token to appear in the cited evidence in source order. It therefore rejects audited negation insertions, out-of-source substitutions, and token reorderings when they violate this contract.  Fixed- and calibrated-coverage baselines remain explicit ablations of this ordered rule.

Activation is a state transition rather than an answer-time score.  If the predicate passes, the store creates $z$ with parent $A_t(\kappa)$ when present and null otherwise, then atomically updates the pointer.  Otherwise persistent state is unchanged.  Accepted proposals become active independently of any future question.  Each decision records the normalized evidence, predicate components, parent version, and resulting status so that admission can be replayed without evaluator data.

\subsection{Conflict-conditioned version resolution}

The deterministic trigger greedily canonicalizes keys in chronology order when normalized subject--relation Jaccard is at least $.60$ or keys match exactly.  A conflict requires different source IDs and values that are neither exact nor Jaccard-equivalent at $.80$; every qualifying pair activates the governed answer route.  Gold answers and update labels are absent.  In the natural runtime, the system uses Dense raw chunks when no qualifying conflict is detected and governed persistent versions otherwise.  Fallback never reverses an accepted write or counts as rollback.

For the controlled temporal protocol, the \emph{Temporal Resolver} groups conflicting candidates by key and applies the declared recency-as-truth contract by selecting
\begin{equation}
\label{eq:temporal-resolve}
\operatorname{Resolve}(C_\kappa)=\arg\max_{q\in C_\kappa}\nu(q).
\end{equation}
Equation~\ref{eq:temporal-resolve} is explicitly scoped to settings in which chronology defines the version visible to the answer model.  The resolver does not establish semantic truth; it only applies the chronology rule defined by the benchmark or application.
Chronology ties are broken deterministically by source order and proposal ID.

\subsection{Durable audit and complete-state recovery}

For a controlled transaction $\tau$, let $S_0$ be the pre-fault persistent state and $A_0=\Pi_{\rm app}(S_0)$ the active-pointer snapshot stored in its recovery intent. Let $I(S)$ denote the conjunction of pointer/version existence, fact-key consistency, active status, and event replay. The audited condition requires both application-state equality and $I(S)$.  The controller first persists $A_0$ and a pending intent, executes the audited update, closes the writer, and reopens the database as $S_1$.  A counted rollback satisfies
\begin{equation}
\label{eq:rollback-contract}
\begin{aligned}
\operatorname{RB}(\tau)\iff{}&
\operatorname{PersistIntent}(A_0)\land\operatorname{Reopen}(S_1)\\
&\land\bigl([\Pi_{\rm app}(S_1)\neq A_0]\lor\neg I(S_1)\bigr)\\
&\land\operatorname{Restore}(A_0)\land\operatorname{Reopen}(S_2)\\
&\land I(S_2)\land\Pi_{\rm app}(S_2)=A_0.
\end{aligned}
\end{equation}
The implementation uses SQLite WAL with \texttt{synchronous=FULL}.  Versions, active pointers, state events, and recovery intents are separate durable tables. The evaluated recovery contract begins when an external detector invokes the controller, isolating durable restoration from deployment-specific anomaly detection.

Snapshot recovery does not assume that one intended logical key identifies the complete physical write set.  One compensating transaction restores the saved active map, marks displaced versions rolled back, records the repair, and finalizes the intent before obtaining $S_2$.  The write-set oracle instead logs preimages for the injected physical write set; matching its coverage demonstrates complete restoration without that privileged information.

\subsection{Contract implications and complexity}

The definitions yield three scoped implications.  If normalization and source lookup implement the frozen predicate, every activated proposal satisfies Equation~\ref{eq:ordered-support}. Under the declared chronology contract, Equation~\ref{eq:temporal-resolve} selects the chronology-maximal visible version with deterministic tie-breaking. If the recovery intent and $A_0$ remain intact and the controller is invoked, Equation~\ref{eq:rollback-contract} targets the complete saved active map rather than a guessed key.  The fault matrix tests both durability boundaries after process termination and reopen.

For $n$ candidates, pairwise trigger construction costs $O(n^2)$ and post-grouping resolution costs $O(n)$.  Ordered PatchTest is linear in the inspected source and token sequences and makes no verifier call.  Snapshot persistence and restoration cost $O(|A_t|)$ keys in exchange for write-set independence.  These bounds exclude shared extraction, retrieval, and answer calls.

\section{Experiments}

\subsection{Evaluation Overview}

We first evaluate the complete write--reject--retain workflow. We then isolate the three components of the transaction boundary. The source-support audit compares write gates on adversarial proposals. The temporal-resolution study measures conflict handling on MemoryAgentBench FactConsolidation. The persistent-fault audit tests complete-state restoration on benchmark-derived LongMemEval-S and LoCoMo states. The final ablations and stress tests identify the source and robustness of the gains.

These experiments use different statistical units: API cases, source items, fixed benchmark histories, and persisted histories. We report these units separately rather than pooling them. Runtime decisions are label-blind. PatchTest sees only proposal fields and the cited source, while the resolver sees candidate keys, values, source IDs, and chronology. Gold answers, benchmark categories, unsafe labels, and fault classes are joined only after decisions. Experiments cover ten local model configurations and two API configurations. Calibration always precedes confirmation, including item-disjoint threshold selection before the held-out source-support audit.

\subsection{End-to-End API-Endpoint Governance}

\begin{figure}[!t]
\centering
\includegraphics[width=\columnwidth]{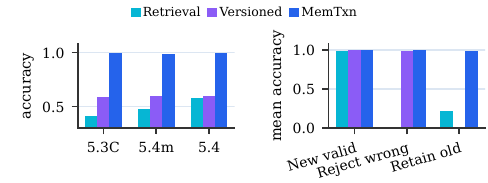}
\twopanellabels{API endpoints}{Task components}
\caption{API-endpoint governance results. Overall workflow accuracy across three endpoint configurations and component-level accuracy for valid writes, wrong-write rejection, and retained tasks.}
\label{fig:closed-boundary}
\end{figure}

\textbf{Workflow and component accuracy.} Figure~\ref{fig:closed-boundary} evaluates 100 controlled cases per endpoint spanning valid writes, unsupported writes, and retained tasks. The compact labels 5.3C, 5.4m, and 5.4 denote the GPT-5.3-Codex-Spark, GPT-5.4-mini, and GPT-5.4 endpoints, respectively. Across these three API-endpoint configurations, each with 300 method rows, MemTxn averages $1.0$ on valid writes and wrong-write rejection and $.992$ on old-task retention. Retrieval retains only $.225$ of old tasks, while Versioned retains none. Endpoint call-error rates remain below $.007$; exact checkpoint revisions were not independently verified. These results show that the same governance boundary admits supported updates, rejects unsupported ones, and preserves prior tasks. The following studies isolate these capabilities and their downstream QA effect.

\begin{table}[!t]
\centering
\small
{\setlength{\tabcolsep}{1.4pt}
\begin{tabular}{lcrrrrrr}
\toprule
Gate & $\theta$ & Acc$\uparrow$ & Bal.$\uparrow$ & Prec.$\uparrow$ & F1$\uparrow$ & FA$\downarrow$ & FR$\downarrow$ \\
\midrule
\textsc{Lexical} & 0.60 & 25.10 & 50.00 & 25.10 & 40.13 & 179 & \textbf{0} \\
\textsc{Calibrated} & 0.95 & 73.22 & 82.12 & 48.39 & 65.22 & 64 & \textbf{0} \\
\textsc{Qwen} & -- & 68.62 & 77.39 & 44.19 & 60.32 & 72 & \underline{3} \\
\textsc{Hybrid} & -- & \underline{78.24} & \underline{83.81} & \underline{53.77} & \underline{68.67} & \underline{49} & \underline{3} \\
\textbf{\textsc{Ordered}} & -- & \textbf{100.00} & \textbf{100.00} & \textbf{100.00} & \textbf{100.00} & \textbf{0} & \textbf{0} \\
\bottomrule
\end{tabular}
}
\caption{Item-disjoint source-support audit. Sixty supported and 179 in-contract hard-negative probes from 12 unseen test items.  Bold/underline mark the best/second-best gate results.}
\label{tab:audits}
\end{table}

\subsection{Source-Support Admission}

\textbf{Admission quality.} This item-disjoint audit is constructed from LongMemEval-S extraction outputs~\cite{longmemeval2024} and tests whether a proposed value may safely become active. The ordered rule is frozen before evaluation. All development item IDs are excluded, and probes remain clustered by source item. Table~\ref{tab:audits} reports 12 unseen test items. They comprise 60 supported originals and 179 hard negatives generated by negation insertion, token reordering, and out-of-source substitution. Ordered PatchTest is the only gate with 100\% accuracy, balanced accuracy, precision, and F1. It accepts every supported original with zero false accepts and zero false rejects. Every comparator makes at least three decision errors, and the strongest alternative still falsely accepts 49 negatives. Thus the result reflects held-out application of the declared rule rather than threshold selection on the test items.

\begin{figure}[!t]
\centering
\includegraphics[width=\columnwidth]{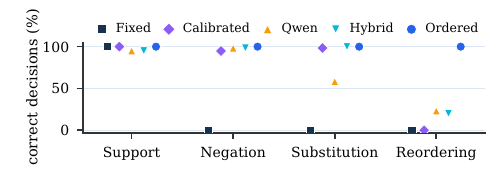}
\caption{Probe-class source-support decisions. Correct-decision rates for five gates on supported originals, negations, substitutions, and reorderings.}
\label{fig:audit-summary}
\end{figure}

\textbf{Error-type breakdown.} Figure~\ref{fig:audit-summary} separates supported originals from the three corruption templates. Ordered PatchTest correctly classifies all 60 originals, 59 negations, 60 substitutions, and 60 reorderings, whereas every comparator fails on at least one category. Thus the ordered rule rejects all 179 audited hard negatives without a verifier call. This controlled result is scoped to the declared ordered lexical-support contract and does not establish semantic-role fidelity or truth.

\subsection{Temporal Resolution on FactConsolidation}

\begin{figure}[!t]
\centering
\includegraphics[width=\columnwidth]{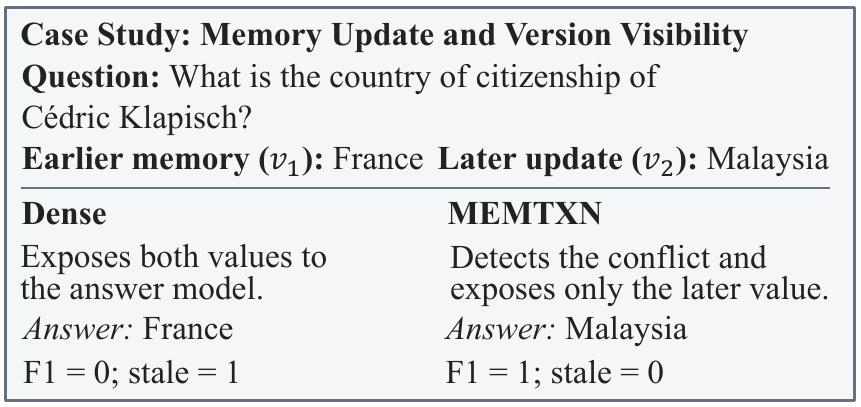}
\caption{Matched-budget FactConsolidation trace with one answer call per
method.}
\label{fig:case-temporal}
\end{figure}

\textbf{Main FactConsolidation results.} MemoryAgentBench FactConsolidation~\cite{memoryagentbench2026} defines the latest conflicting assertion as correct, matching the resolver's chronology contract. All methods use the same prompt and one answer call; MemTxn changes only version visibility. Table~\ref{tab:factconsol} reports aggregate and breakdown metrics for seven methods on five representative models (800 questions each), while Figure~\ref{fig:case-temporal} traces one matched-budget example.

MemTxn leads every model block, improving over Dense by 17.06--24.07 F1 points and reducing conditional stale rate by 15.97--37.6 points. Larger single-hop than multi-hop gains indicate remaining answer-reasoning limits.

\begin{table*}[!t]
\centering
\small
{\setlength{\tabcolsep}{2.9pt}
\renewcommand{\arraystretch}{0.90}
\begin{tabular}{llrrrrrrrrrrr}
\toprule
\textbf{Model} & \textbf{Method} & \multicolumn{4}{c}{\textbf{Overall (\%)}} & \multicolumn{2}{c}{\textbf{Hop (\%)}} & \multicolumn{4}{c}{\textbf{Context length (\%)}} & \textbf{Tok./q$\downarrow$} \\
\cmidrule(lr){3-6}\cmidrule(lr){7-8}\cmidrule(lr){9-12}
 &  & \textbf{F1$\uparrow$} & \textbf{$\Delta$Dense$\uparrow$} & \textbf{SubEM$\uparrow$} & \textbf{Stale$\downarrow$} & \textbf{SH$\uparrow$} & \textbf{MH$\uparrow$} & \textbf{6K$\uparrow$} & \textbf{32K$\uparrow$} & \textbf{64K$\uparrow$} & \textbf{262K$\uparrow$} &  \\
\midrule
\multirow{7}{*}{Qwen2.5-7B} & \textsc{Bm25} & 22.92 & +2.03 & 23.00 & 31.79 & 40.60 & 5.24 & 18.30 & 25.02 & 25.51 & 22.85 & 168 \\
 & \textsc{Dense} & 20.89 & +0.00 & 19.75 & 49.50 & 38.46 & 3.31 & 15.80 & 24.57 & 23.73 & 19.44 & 166 \\
 & \textsc{Mem0} & 23.32 & +2.43 & 23.12 & 43.65 & 41.03 & 5.61 & 18.45 & 26.41 & 24.15 & 24.27 & 166 \\
 & \textsc{A-Mem} & \underline{36.66} & \underline{+15.78} & \underline{36.00} & 25.65 & \underline{65.29} & 8.04 & \underline{38.12} & 35.95 & \underline{37.26} & \underline{35.33} & \underline{165} \\
 & \textsc{Zep} & 33.71 & +12.83 & 32.88 & 15.40 & 56.27 & \underline{11.16} & 34.39 & \underline{38.57} & 32.49 & 29.41 & 167 \\
 & \textsc{LightMem} & 21.10 & +0.22 & 20.00 & 46.47 & 37.86 & 4.34 & 17.03 & 23.27 & 25.68 & 18.43 & \textbf{120} \\
 & \textbf{\textsc{MemTxn}} & \textbf{42.88} & \textbf{+21.99} & \textbf{42.12} & 15.11 & \textbf{73.65} & \textbf{12.10} & \textbf{47.39} & \textbf{41.35} & \textbf{45.38} & \textbf{37.40} & 167 \\
\midrule
\multirow{7}{*}{Qwen3.5-9B} & \textsc{Bm25} & 25.54 & +4.16 & 26.88 & 28.15 & 46.71 & 4.37 & 20.76 & 27.65 & 25.66 & 28.08 & 163 \\
 & \textsc{Dense} & 21.38 & +0.00 & 21.62 & 47.19 & 39.23 & 3.52 & 16.62 & 25.52 & 23.74 & 19.65 & 161 \\
 & \textsc{Mem0} & 27.25 & +5.87 & 28.88 & 39.56 & 48.70 & 5.80 & 24.26 & 30.63 & 27.62 & 26.48 & \underline{160} \\
 & \textsc{A-Mem} & \underline{40.66} & \underline{+19.28} & \underline{41.38} & 23.63 & \underline{72.04} & 9.28 & \underline{44.59} & \underline{39.10} & \underline{38.66} & \textbf{40.30} & \underline{160} \\
 & \textsc{Zep} & 34.31 & +12.93 & 33.88 & 10.36 & 56.24 & \underline{12.37} & 35.72 & 34.83 & 35.70 & 30.98 & 161 \\
 & \textsc{LightMem} & 22.93 & +1.55 & 23.25 & 44.03 & 40.15 & 5.72 & 19.00 & 27.33 & 24.83 & 20.57 & \textbf{111} \\
 & \textbf{\textsc{MemTxn}} & \textbf{45.45} & \textbf{+24.07} & \textbf{45.50} & 9.64 & \textbf{77.05} & \textbf{13.85} & \textbf{50.03} & \textbf{43.33} & \textbf{48.62} & \underline{39.80} & 161 \\
\midrule
\multirow{7}{*}{Qwen3.5-35B-A3B} & \textsc{Bm25} & 28.37 & +1.77 & 35.62 & 21.15 & 51.63 & 5.10 & 24.05 & 29.20 & 31.21 & 29.00 & 164 \\
 & \textsc{Dense} & 26.59 & +0.00 & 31.75 & 40.14 & 48.83 & 4.36 & 22.46 & 27.93 & 30.18 & 25.80 & 160 \\
 & \textsc{Mem0} & 29.38 & +2.78 & 35.62 & 32.26 & 52.75 & 6.00 & 27.65 & 31.73 & 29.93 & 28.20 & 159 \\
 & \textsc{A-Mem} & \underline{41.50} & \underline{+14.91} & \underline{44.12} & 22.19 & \underline{72.80} & 10.20 & \underline{46.76} & \underline{39.15} & \underline{39.71} & \underline{40.38} & \underline{158} \\
 & \textsc{Zep} & 32.01 & +5.42 & 31.87 & 6.62 & 51.26 & \underline{12.77} & 32.90 & 33.98 & 33.23 & 27.94 & 161 \\
 & \textsc{LightMem} & 26.46 & -0.14 & 32.12 & 37.41 & 47.96 & 4.96 & 22.22 & 29.87 & 28.55 & 25.19 & \textbf{111} \\
 & \textbf{\textsc{MemTxn}} & \textbf{47.96} & \textbf{+21.37} & \textbf{50.38} & 9.64 & \textbf{80.66} & \textbf{15.27} & \textbf{53.94} & \textbf{44.71} & \textbf{50.30} & \textbf{42.90} & 160 \\
\midrule
\multirow{7}{*}{Llama-3.2-3B} & \textsc{Bm25} & 25.43 & +1.27 & 24.38 & 29.83 & 46.75 & 4.11 & 21.92 & 27.39 & 28.26 & 24.16 & 175 \\
 & \textsc{Dense} & 24.16 & +0.00 & 22.88 & 46.04 & 43.47 & 4.85 & 19.64 & 24.25 & 28.67 & 24.07 & 173 \\
 & \textsc{Mem0} & 26.54 & +2.38 & 25.12 & 42.04 & 45.90 & 7.19 & 24.85 & 27.94 & 27.48 & 25.90 & \underline{172} \\
 & \textsc{A-Mem} & \underline{39.38} & \underline{+15.23} & \underline{38.25} & 25.36 & \underline{69.55} & 9.22 & \underline{43.53} & \underline{37.13} & \underline{37.73} & \underline{39.14} & \underline{172} \\
 & \textsc{Zep} & 32.90 & +8.74 & 32.38 & 11.65 & 55.00 & \underline{10.81} & 32.31 & 34.62 & 34.81 & 29.88 & 174 \\
 & \textsc{LightMem} & 23.96 & -0.20 & 22.88 & 42.30 & 43.21 & 4.70 & 20.28 & 25.29 & 26.98 & 23.27 & \textbf{128} \\
 & \textbf{\textsc{MemTxn}} & \textbf{44.25} & \textbf{+20.09} & \textbf{43.38} & 11.08 & \textbf{76.05} & \textbf{12.44} & \textbf{47.23} & \textbf{40.77} & \textbf{47.68} & \textbf{41.31} & 174 \\
\midrule
\multirow{7}{*}{GPT-5.4} & \textsc{Bm25} & 29.31 & -2.94 & 40.75 & 15.27 & 55.60 & 3.02 & 26.13 & 31.02 & 31.66 & 28.44 & 539 \\
 & \textsc{Dense} & 32.25 & +0.00 & 46.12 & 22.59 & 58.98 & 5.52 & 30.38 & 33.12 & 34.51 & 30.99 & 534 \\
 & \textsc{Mem0} & 33.12 & +0.87 & 46.00 & 21.46 & 57.88 & 8.36 & 31.21 & 36.51 & 32.04 & 32.72 & 535 \\
 & \textsc{A-Mem} & \underline{42.37} & \underline{+10.12} & \underline{48.12} & 17.15 & \underline{75.63} & 9.11 & \underline{46.88} & \underline{40.42} & \underline{40.20} & \underline{41.98} & \underline{514} \\
 & \textsc{Zep} & 31.27 & -0.98 & 32.12 & 5.18 & 50.63 & \underline{11.92} & 32.83 & 33.62 & 31.48 & 27.17 & 529 \\
 & \textsc{LightMem} & 31.62 & -0.63 & 43.88 & 23.60 & 58.27 & 4.97 & 30.65 & 32.21 & 31.82 & 31.80 & \textbf{493} \\
 & \textbf{\textsc{MemTxn}} & \textbf{49.31} & \textbf{+17.06} & \textbf{55.38} & 6.62 & \textbf{83.95} & \textbf{14.66} & \textbf{55.88} & \textbf{46.90} & \textbf{52.05} & \textbf{42.39} & 516 \\
\bottomrule
\end{tabular}
}
\caption{MemoryAgentBench FactConsolidation results. Five representative model settings, seven methods, and 800 QA rows per model; $\Delta$Dense is the absolute F1 difference in percentage points. Bold/underline mark best/second-best comparable values; conditional stale rates are unranked.}
\label{tab:factconsol}
\end{table*}

\textbf{Matched top-8 control.} With both candidate pools fixed at eight, Table~\ref{tab:matched-top8} shows MemTxn gains of 15.01--22.93 F1 points across six models (800 questions each), with every model-level 95\% confidence interval above zero. The pooled gain is +19.51 points over 4,800 pairs; stale rate is omitted because the reused Dense eligibility pool used retrieval depth 24.

\begin{table}[!t]
\centering
\small
{\setlength{\tabcolsep}{1.5pt}
\begin{tabular}{@{}lrrrrr@{}}
\toprule
Model & \multicolumn{2}{c}{F1 (\%)} & $\Delta$ & 95\% CI & W/T/L \\
\cmidrule(lr){2-3}
 & D & M & & & \\
\midrule
Q2.5-7B & 20.89 & \textbf{41.94} & +21.05 & $[18.10,24.05]$ & 187/601/12 \\
Q3.5-9B & 21.44 & \textbf{44.37} & +22.93 & $[19.96,25.99]$ & 205/584/11 \\
Q3.5-35B & 26.86 & \textbf{46.41} & +19.55 & $[16.79,22.39]$ & 201/585/14 \\
L3.2-3B & 24.14 & \textbf{44.35} & +20.21 & $[17.26,23.20]$ & 178/610/12 \\
GPT-5.4m & 30.15 & \textbf{48.44} & +18.29 & $[15.77,20.87]$ & 220/564/16 \\
GPT-5.4 & 32.80 & \textbf{47.81} & +15.01 & $[12.83,17.27]$ & 210/573/17 \\
\midrule
Pooled & 26.05 & \textbf{45.55} & +19.51 & $[18.38,20.63]$ & 1201/3517/82 \\
\bottomrule
\end{tabular}
}
\caption{Matched top-8 F1 over 800 questions per model (4,800 pooled pairs); bold marks the higher F1.}
\label{tab:matched-top8}
\end{table}

\begin{figure}[!t]
\centering
\includegraphics[width=\columnwidth]{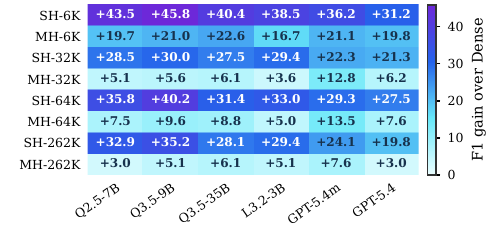}
\caption{MemTxn--Dense F1 gains across models and hop/context-length settings.}
\label{fig:temporal-summary}
\end{figure}

\textbf{Cross-model gains.} Figure~\ref{fig:temporal-summary} summarizes 48 default-protocol cells across six models. All gains are positive and consistently larger for single-hop than multi-hop questions; default- and matched-pool magnitudes remain incomparable.

\subsection{Complete-State Recovery}

\textbf{Recovery scope and cost.} The audit applies four persistent fault classes to 58 LongMemEval-S histories and 10 LoCoMo conversations, spanning 5--64 active and 2--16 affected keys. Passing requires complete pre-fault restoration under status, pointer/version/fact-key, event-replay, and second-reopen checks.

Table~\ref{tab:recovery-main} reports the hardest setting, with 64 active keys and 16 affected keys. MemTxn and the write-set oracle have zero failures over 272 runs, whereas the single-key baseline fails all 136 partial-commit and event-divergence runs. Thus MemTxn matches the privileged oracle's coverage without receiving the physical write set. No design falsely rolls back a valid commit.

\begin{table}[!t]
\centering
\small
{\setlength{\tabcolsep}{1.3pt}
\begin{tabular}{lllllrr}
\toprule
Source & Design & Record & W-set? & \shortstack{Obs.\\scope} & \shortstack{P/E\\kB} & \shortstack{P/E\\att. ms} \\
\midrule
\multirow{3}{*}{LME-S} & \textsc{Single} & One key & No & A/W & 0.08 & 20.95 \\
& \textsc{W-set} & Actual W-set & Yes & A/W/P/E & 0.91 & 20.74 \\
& \textbf{\textsc{MemTxn}} & Active map & No & A/W/P/E & 3.62 & 22.49 \\
\midrule
\multirow{3}{*}{LoCoMo} & \textsc{Single} & One key & No & A/W & 0.07 & 20.21 \\
& \textsc{W-set} & Actual W-set & Yes & A/W/P/E & 0.66 & 19.61 \\
& \textbf{\textsc{MemTxn}} & Active map & No & A/W/P/E & 2.63 & 20.08 \\
\bottomrule
\end{tabular}
}
\caption{Recovery scope and cost; \textsc{W-set} is the oracle.}
\label{tab:recovery-main}
\end{table}

\subsection{Ablations and Stress Tests}

\begin{figure}[!t]
\centering
\includegraphics[width=\columnwidth]{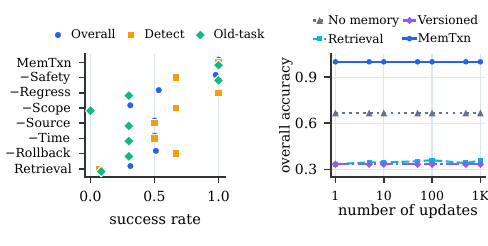}
\twopanellabels{Component ablation}{Accuracy vs. update count}
\caption{Component ablations and update-scale accuracy.}
\label{fig:mechanism-summary}
\end{figure}

\textbf{Mechanism ablations and update scale.} Figure~\ref{fig:mechanism-summary} shows that removing any governance component lowers at least one of overall success, detection, or old-task retention. Only full MemTxn reaches $1.0$ on all three component targets and maintains it through 1,000 updates.

\textbf{Gate reliability.} Table~\ref{tab:accounting-gate} shows smooth degradation from 100.00\% accuracy at $r=100\%$ to 62.47\% at $r=50\%$, where the gain over the strongest non-MemTxn control remains 11.35 points. The monotone margin indicates graceful failure rather than a sharp collapse under imperfect gating.

\begin{table}[!t]
\centering
\small
{\setlength{\tabcolsep}{1.2pt}
\begin{tabular}{crrrrrr}
\toprule
$r$ (\%) & Overall$\uparrow$ & $\Delta$Best$\uparrow$ & Detect$\uparrow$ & Recover$\uparrow$ & Unsafe$\downarrow$ & Forget$\downarrow$ \\
\midrule
100 & \textbf{100.00} & \textbf{+48.89} & \textbf{100.00} & \textbf{100.00} & \textbf{0.00} & \textbf{0.00} \\
90  & \underline{92.63} & \underline{+41.51} & \underline{96.70} & \underline{94.35} & \underline{3.30} & \underline{6.76} \\
80  & 84.99 & +33.88 & 93.30 & 87.76 & 6.70 & 13.60 \\
70  & 77.75 & +26.64 & 89.50 & 81.55 & 10.50 & 20.63 \\
60  & 70.03 & +18.91 & 86.70 & 73.70 & 13.30 & 27.88 \\
50  & 62.47 & +11.35 & 83.60 & 64.98 & 16.40 & 34.74 \\
\bottomrule
\end{tabular}
}
\caption{Gate-reliability stress; $\Delta$Best is the gain over the strongest control at the same $r$.}
\label{tab:accounting-gate}
\end{table}

\begin{figure}[!t]
\centering
\includegraphics[width=\columnwidth]{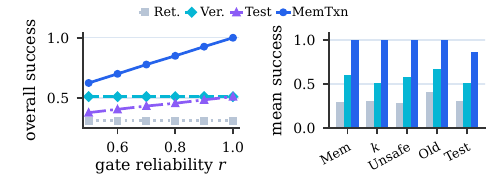}
\twopanellabels{Gate reliability}{Stress axes}
\caption{Robustness to gate reliability and sensitivity axes.}
\label{fig:robustness-summary}
\end{figure}

\textbf{Robustness axes.} Figure~\ref{fig:robustness-summary} shows smooth degradation as gate reliability falls and higher MemTxn accuracy than retrieval and versioned memory on all five retained sensitivity axes. The advantage persists across memory size, retrieval depth, unsafe-update rate, old-task mix, and test composition, rather than depending on one stress dimension.

\section{Discussion and Limitations}

The results support source-constrained admission, chronology-based visibility, and complete-state recovery under the declared contracts, but not semantic truth or robustness to concurrent or repeated faults, intent corruption, or physical loss. Coverage under naturally occurring update triggers remains outside the controlled evaluation. Production deployment additionally requires consent, retention, encryption, access control, and audit minimization.

\section{Conclusion}

We presented \system{}, an answer-model-external transaction boundary for writable agent memory. It separates source-supported update admission, chronology-conditioned answer selection, and invariant-checked complete-state recovery. Ordered PatchTest accepted all 60 supported originals and rejected all 179 template-generated hard negatives. \system{} also restored the complete declared active map across persistent multi-key faults without the actual physical write set. On MemoryAgentBench FactConsolidation, \system{} achieved the highest average F1 across all twelve answer-model configurations. It improved F1 over Dense by $17.06$--$24.07$ points in five representative settings. The six-model matched-top-8 control retained gains of $15.01$--$22.93$ points. Together, these results showed that auditable commit decisions, explicit version visibility, and durable recovery could jointly improve the reliability of long-term agent memory under declared contracts. Future work will extend \system{} to handle concurrent and repeated faults, broaden natural-input governance coverage, and recover from physical storage loss.

\bibliography{references}

\end{document}